\documentclass[manuscript,screen,sigconf]{acmart}

\usepackage{xcolor}
\usepackage{algorithm} 
\usepackage{algpseudocode} 
\usepackage{svg}
\usepackage{booktabs}

\makeatletter
\def\algbackskip{\hskip-\ALG@thistlm}
\makeatother

\AtBeginDocument{%
  \providecommand\BibTeX{{%
    \normalfont B\kern-0.5em{\scshape i\kern-0.25em b}\kern-0.8em\TeX}}}

\copyrightyear{2024}
\acmYear{2024}
\setcopyright{rightsretained}
\acmConference[GECCO '24]{Genetic and Evolutionary Computation Conference}{July 14--18, 2024}{Melbourne, VIC, Australia}
\acmBooktitle{Genetic and Evolutionary Computation Conference (GECCO '24), July 14--18, 2024, Melbourne, VIC, Australia}
\acmDOI{10.1145/3638529.3654125}
\acmISBN{979-8-4007-0494-9/24/07}

\begin{document}

\title[Temporal surrogate fitness landscapes]{Temporal True and Surrogate Fitness Landscape Analysis \\ for Expensive Bi-Objective Optimisation}

\author{Cedric J. Rodriguez}
\email{c.j.rodriguez@lumc.nl}
\affiliation{%
  \institution{Leiden University Medical Center}
  \streetaddress{P.O. Box 1212}
  \city{Leiden}
  \country{The Netherlands}
  \postcode{43017-6221}
}


\author{Sarah L. Thomson}
\email{s.thomson4@napier.ac.uk}
\affiliation{%
  \institution{ Edinburgh Napier University }
  \streetaddress{Sighthill Court}
  \city{Edinburgh}
  \country{Scotland}
  \postcode{EH11 4BN}
}

\author{Tanja Alderliesten}
\email{t.alderliesten@lumc.nl}
\affiliation{%
  \institution{Leiden University Medical Center}
  \streetaddress{P.O. Box 1212}
  \city{Leiden}
  \country{The Netherlands}
  \postcode{43017-6221}
}

\author{Peter A.N. Bosman}
\email{peter.bosman@cwi.nl}
\affiliation{%
  \institution{Centrum Wiskunde \& Informatica}
  \streetaddress{P.O. Box 1212}
  \city{Amsterdam}
  \country{The Netherlands}
  \postcode{43017-6221}
  }

\renewcommand{\shortauthors}{Rodriguez, et al.}

\begin{abstract}
Many real-world problems have expensive-to-compute fitness functions and are multi-objective in nature. Surrogate-assisted evolutionary algorithms are often used to tackle such problems. Despite this, literature about analysing the fitness landscapes induced by surrogate models is limited, and even non-existent for multi-objective problems. This study addresses this critical gap by comparing landscapes of the true fitness function with those of surrogate models for multi-objective functions. Moreover, it does so temporally by examining landscape features at different points in time during optimisation, in the vicinity of the population at that point in time. We consider the BBOB bi-objective benchmark functions in our experiments. The results of the fitness landscape analysis reveals significant differences between true and surrogate features at different time points during optimisation. Despite these differences, the true and surrogate landscape features still show high correlations between each other. Furthermore, this study identifies which landscape features are related to search and demonstrates that both surrogate and true landscape features are capable of predicting algorithm performance. These findings indicate that temporal analysis of the landscape features may help to facilitate the design of surrogate switching approaches to improve performance in multi-objective optimisation. 
\end{abstract}



\maketitle

\section{Introduction} 
\label{sec:intro}

In real-world optimisation problems, fitness evaluations can be computationally expensive --- ranging from minutes to hours \cite{li2022evolutionary} --- and multi-objective (MO) in nature. This greatly limits the total amount of possible fitness evaluations to typically a few hundred or thousand. For such expensive-to-evaluate problems, surrogate-assisted evolutionary algorithms (SA-EAs) have been introduced \cite{he2023review}. SA-EAs reduce the required (true) expensive problem evaluations by substituting the true fitness function during parts of the search, with a quick-to-evaluate model (surrogate) of the problem built on a set of true evaluated solutions. In one scenario that is akin to Bayesion optimisation, the SA-EA iteratively optimises the surrogate function, selects one or more surrogate optimised solutions and evaluates using the true function, and updates the surrogate model based on the new set of true evaluated solutions. Because this surrogate is built on a very limited number of solutions, there are likely discrepancies between the true fitness landscape and the surrogate fitness landscape. Therefore, the performance of the SA-EAs is not only dependent on the true fitness function but also on the surrogate fitness function. In this work, we analyse these potential discrepancies between the true and surrogate landscape to better understand the relation between the optimisation performance, the surrogate, and the true fitness landscape during the optimisation.\

To properly study potential discrepancies between the true and surrogate fitness landscapes, we need to characterise fitness landscapes during different points of the optimisation because the surrogate landscape may constantly shift due to the surrogate model being updated. Fitness landscape analysis (FLA) can do that, providing insight about the interplay between optimisation algorithms and different fitness landscape. The nature of FLA depends on whether the search space is discrete or continuous, and on how many objectives are formulated. We consider in this work continuous bi-objective problems, and compute features of the true fitness landscape and also surrogate landscapes. In the continuous optimisation domain, many FLA works have focused on algorithm-agnostic (static) approaches, such as Latin hypercube sampling (LHS) \cite{munoz2014exploratory,mersmann2011exploratory,bischl2012algorithm,liefooghe2021landscape} and random walks \cite{lang2020distributed,jana2018continuous,lang2019robustness}. The nature of the sampling means that the computed landscape metrics may not be closely related to the  landscape that is observed through fitness evaluations performed during an actual optimisation run. In a survey on landscape analysis, Malan and Engelbrecht referred to this as \emph{search independence} \cite{malan2013survey}. There are also several works which use search-dependent approaches: one of these is continuous-space local optima networks (LONs) \cite{adair2019local,tomassini2022local}. LONs, although valuable, amalgamate information from across the evolutionary timespan into a single mathematical object (the LON). There is also the multi-objective equivalent, Pareto LONs (PLONs) \cite{liefooghe2018pareto}. PLONs have several associated features which have been linked to search performance; however, we do not use them here because they do not fully describe the landscape as observed by an algorithm during optimisation.

Different to the static FLAs, Jankovic and Doerr considered the evolution of exploratory landscape analysis features in a single-objective context \cite{jankovic2019adaptive}, finding that static and temporal landscapes do not match. Several works have studied the notion of \emph{trajectory features} (for single-objective problems): these allow an algorithm to begin optimising while logging its trajectory, then computing features from it to understand the problem or make algorithmic decisions \cite{jankovic2021towards,jankovic2022trajectory,kostovska2022per,cenikj2023dynamorep,vermetten2023switch}. These contributions are related to ours because they leverage intermediate landscape information yet focus on single-objective problems. We agree with the view that studying temporal features associated with using an optimisation algorithm, or considering the evolution of them over time may be valuable. Very recently, Alsouly \emph{et al.} studied the evolution of landscape features over time for a set of constrained multi-objective problems \cite{alsouly2023dynamic}, finding that the landscape shifts across the course of evolution, and that predicting algorithm performance is possible with temporal features. In that work, several of the features related to the constrained nature of the problems. Two previous studies \cite{pitra2019landscape,pitra2022landscape} have computed metrics at different points in evolution for a surrogate-guided evolutionary algorithm, but these were joined together for analysis and the change over time was not separated out and studied. In that case, the sample points used for landscape analysis were based on the true fitness function, and the problems were single-objective; the authors found that several metrics are linked to surrogate-assisted algorithm performance.

Aside from the mentioned works, there are very few articles considering surrogate fitness landscapes. Werth \emph{et al.} explore this direction \cite{werth2020surrogate}, comparing surrogate landscapes with those of the true fitness function for a small number of single-objective problems. Harrison \emph{et al.} analyse the surrogate landscape for single-objective parameter configuration \cite{harrison2023surrogate}, but do not compare with the true landscape. Generally, there is a lack of literature analysing surrogate landscapes. In the context of multi-objective optimisation, the literature for this is --- to the best of our knowledge --- non-existent.

 The present work is the first to conduct analysis of multi-objective surrogate landscapes. Additionally, we compare temporal landscape features of the true fitness landscape and the constantly-shifting surrogate fitness landscape, emphasizing the locally encountered landscape and how it changes over time. Although there has been an initial work studying multi-objective features across evolution \cite{alsouly2023dynamic}, this was for constrained problems and did not consider surrogates. Additionally, we consider here a different set of functions and features and argue that this is an under-explored avenue. Another contribution is that we demonstrate the potential associated with building algorithm performance prediction models using features of \emph{both} the surrogate and true landscape. The rest of this paper is structured as follows: Section \ref{sec:relawork} will introduce the surrogate models, EAs, and FLA methods, and dimensionality reduction methods that will be considered in this work. Sections \ref{methods} and \ref{ExperimentalSetup} will describe our main methodology and experimental setup, respectively. In Section \ref{results}, the experimental results will be presented. Finally, Section \ref{sec:conclusions} consists of the conclusions of the work.

\section{Background} 
\label{sec:relawork}

\subsection{MO Surrogate-assisted EA}
\label{bg:SUR}
Figure \ref{fig:SAEA-METHOD}(a), describes the main process of the type of MO SA-EA that we consider in this work. During the initialization phase, the SA-EA samples a set of solutions using LHS, evaluates them using the true fitness function, and builds the initial surrogate model. After the initialization, the SA-EA is composed of an inner cycle and an outer cycle. The inner cycle consists of typical generational processes of an EA. This cycle initializes with a population containing all previously evaluated solutions using the true fitness function. After the application of the selection and variational operator, offspring solutions are evaluated using the quick-to-evaluate surrogate model instead of the expensive-to-evaluate problem. The cycle terminates as soon as the maximum number of surrogate evaluations have been performed. After the termination of the generational cycle, the surrogate optimisation cycle starts. All surrogate-evaluated solutions are then considered as potential candidate solutions for evaluation with the true fitness function. In the multi-objective context considered for this work, the selection process consists of calculating the domination ranks for all solutions based on their surrogate fitness values. A pre-selection is made by excluding 50\% of the solutions with the largest ranks. Thereafter, a set of solutions is randomly selected from the pre-selection for the true evaluation. The true evaluated solutions are then logged and, subsequently, a surrogate model is updated based on this new set of solutions. This SA-EA is described more formally in Algorithm \ref{alg:SA-EA}.

\begin{algorithm}
	\caption{MO SA-EA}
	\raggedright
	\hspace*{\algorithmicindent} \textbf{Input:} $N$ = Number of initial solutions; $FE_{max}^{true}$ = maximum number of true fitness evaluations; $FE_{max}^{surrogate}$ = maximum number of surrogate evaluations in each surrogate optimisation cycle; $\mu$ = number of new selected solutions after each surrogate optimisation cycle; \\
    \hspace*{\algorithmicindent} \textbf{Output:} $A$  = Archive of all non-dominated solutions
	\begin{algorithmic}[1]
        \State Start first surrogate optimisation cycle and set true function evaluation $FE^{true}$ to 0
	    \State Initialize population $P$ of size $N$ using random sampling and evaluate solutions using the true fitness function
        \State Add initial population $P$ to archive $A$
		\While {$ FE < FE_{max}^{true}$}
            \State Build surrogate model $M$ based on $A$
            \State Optimize $M$ using EA with $FE_{max}^{surrogate}$
		    \State Select $\mu$ solutions from all surrogate evaluated solutions to evaluate with true fitness function ($FE^{true} = FE^{true} + \mu$)
		    \State Add solutions to the archive
		\EndWhile
        \State Return $A$
	\end{algorithmic} 
    \label{alg:SA-EA}
\end{algorithm}

\subsection{Surrogate models}
\label{bg:models}
\subsubsection{Exact interpolators}
In previous expensive multi-objective optimisation algorithms, several point interpolation methods have been considered, such as inverse distance weighting (IDW) \cite{shepard1968two}. \emph{IDW} is one of the most straightforward and widely adopted \cite{li2011review} point interpolation methods, defined as follows:
\vspace{-1em}

\begin{equation}
    \hat{y}^{\mbox{\scriptsize{\emph{idw}}}}(x) = \sum_{i=1}^n w_{i}(x) y_{i}
    \label{eq:IDW}
\end{equation}

where $\hat{y}^{\mbox{\scriptsize{\emph{idw}}}}$ is the surrogate fitness value of a solution with an unknown true fitness value, $y_i$ is the true fitness value of the $ith$ solution in an archive $A_i$ containing $n$ solutions, and 
\vspace{-1em}

\begin{equation}
    w_{i}(x) = \frac{d(x, A_i)^{-1}}{\Sigma_{j=1}^n d(x, A_j)^-1}
    \label{eq:IDW_weights}
\end{equation}

where $d$ is the Euclidean distance and $x$ the unknown solution and the $ith$ solution. A downside of using \emph{IDW} is that all surrogate values of the solutions outside the parameter bounds of the known solutions approach the average fitness of the known solutions. To overcome this limitation, inverse distance weighting regression (IDWR) has recently been developed \cite{emmendorfer2020novel} where the surrogate fitness values of solutions outside the parameter bounds follow the trend line of the fitness value of all known solutions. The surrogate value using the \emph{IDWR} interpolation method is computed as follows
\vspace{-1em}

\begin{equation}
    \hat{y}^{\mbox{\scriptsize{\emph{idwr}}}}(x) = \hat{y}^{\mbox{\scriptsize{\emph{idw}}}}(x) + n\frac{\Sigma_{i=1}^n y_i - n \hat{y}^{\mbox{\scriptsize{\emph{idw}}}}(x)}{n^2 - \Sigma_{i=1}^n d(x, A_i) \Sigma_{i=1}^n d(x, A_i)^{-1}}
    \label{eq:IDWR}
\end{equation}

\subsubsection{Linear regression}
Another surrogate model is linear regression. To preserve the local structure of the fitness landscape, the surrogate is built on the k-nearest neighbours (KNN) of the solution with unknown true fitness value.  Linear regression is defined as:
\vspace{-1em}

\begin{equation}
    \hat{y}^{\mbox{\scriptsize{\emph{lr-knn}}}}(x) = \beta_0 + \sum_{q=1}^D \beta_{q} x_{q}
\end{equation}

where $ \hat{y}^{\mbox{\scriptsize{\emph{lr-knn}}}}(x)$ is the surrogate fitness value, $\beta_q$ are the coefficient (slope) in each problem dimension, $D$ is the number of problem dimensions and $x_{q}$ are the decision variables. 

\subsection{Reference vector guided EA}
The optimizer for the SA-EA in this work is a reference vector guided evolutionary algorithm (RVEA) \cite{cheng2016reference}. The initial population is generated by uniformly sampling solutions in the initialization range of the multi-objective problem. As long as the maximum number of function evaluations is not exceeded, simulated binary crossover \cite{deb1995simulated} followed by polynomial mutation \cite{deb1996combined} is carried out for generating an offspring population from the parent population. This method for generating offspring solutions is equivalent to other multi-objective EAs such as NSGA-III \cite{deb2013evolutionary}. After combining the parent and offspring population, the parent population for the next generation is selected using a reference vector guided procedure called the angle penalized distance.


\subsubsection{Selection using angle penalized distance}
Angle penalized distance \cite{cheng2016reference} consists of four mechanisms: generation of reference vectors, assignment of each solution to one of the reference vectors, selection of a solution per reference vector, and the adaption of reference vectors. With RVEA, the objective space is divided into discrete partitions using reference vectors. The initial generation of these uniform reference vectors is done by firstly generating a set of uniformly distributed reference points. In the next step, each solution in the population is assigned to the reference vector with the smallest angle between the objective vector (the vector between the objective values and the origin) and the reference vector. This partitions the population into several subpopulations, where each subpopulation is associated with a particular reference vector. In the third step, one solution is selected per subpopulation. The selection criterion consists of two subcriteria. The first subcriterion (convergence) is measuring the magnitude of the objective vector, where a smaller objective vector means a better solution. The second subcriterion (divergence) is the angle between the objective vector and the reference vector, where a larger angle means more divergence. These two subcriteria are combined into a single criterion: the angle penalized distance, where the parameter $\alpha$ specifies the balance between the subcriteria. 

\section{Methodology} 
\label{methods}
\begin{figure*}
  \includegraphics[width=0.9\textwidth]{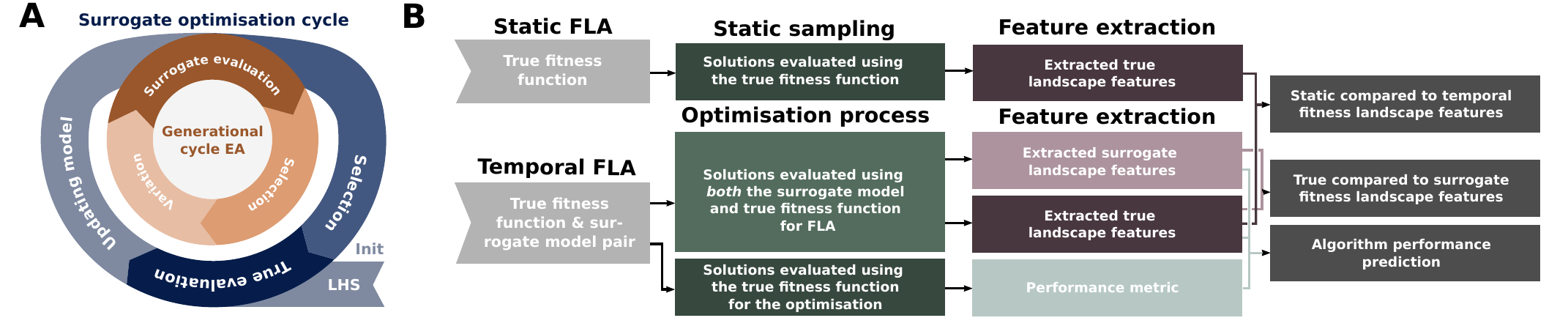}
  \caption{A) Visualisation of the main process of the type of SA-EA that we consider in this work where the (orange) inner cycle represents a generational cycle of an EA with a selection and variational operator. Within this cycle, the solutions are evaluated using a surrogate model. After termination of the generational cycles, the (blue) outer cycle starts, where potentially good surrogate evaluated solutions are selected and evaluated with the true fitness function. B) The general methodology is composed of a static FLA and a temporal FLA where the static FLA utilizes a static sampling strategy while the temporal FLA logs the solutions during the optimisation. A feature extraction process then extracts the true and surrogate landscape features. Furthermore, the performance metric is calculated, which is used for algorithm performance prediction.}
  \label{fig:SAEA-METHOD}
\end{figure*}

The main processes in the methodology are presented in Figure \ref{fig:SAEA-METHOD}(b). The methodology is composed of a static FLA and a temporal FLA. For the static FLA, $200D$ solutions are sampled using LHS, where $D$ is the number of problem dimensions. These solutions are used to extract the true fitness landscape features. In the temporal FLA, for every problem and surrogate model pair, we independently run an optimisation process using an SA-EA. Independent optimisation runs driven by each surrogate model are required because different models might guide the EA to different parts of the search space and encounter different fitness landscapes. During each optimisation, solutions are logged alongside both the surrogate fitness and also the true fitness. From these samples, landscape features and optimisation metrics are extracted for the final analyses. 

\subsection{Sampling during optimisation}
During 15 repeats of the optimisation, the locally encountered true and surrogate fitness landscape is sampled at each time point in the following way: from the current population, we repeatedly apply the selection and variational operators until 2000 evaluations have been reached; we thereby sample the vicinity around the population. From this sample, duplicates are removed, meaning that the sample can be slightly less than 2000. Two different sets of solutions are thus logged after every surrogate optimisation cycle:
\begin{enumerate}
    \item Sample of solutions that have been evaluated by the SA-EA within the current surrogate optimisation cycle. The solutions are also evaluated using the true fitness function to acquire the true fitness values.
    \item The true evaluated solutions for the optimisation.
\end{enumerate}

\subsection{Feature extraction process}
In the feature extraction process, the true and surrogate landscape features are extracted using the FLA method from Liefooghe et al. \cite{liefooghe2021landscape} designed for continuous multi-objective problems. From the 49 features, we removed any features which displayed missing values in any of the samples. Missing values can occur for some of the features depending on the nature of the sample; for example, if there is only one non-dominated point in a sample, then there will be missing values for the metrics relating to distances between non-dominated points. Additionally, we remove a group of features which are associated with hypervolume; because hypervolume is the measurement we will be predicting, it seemed appropriate to exclude them. The resulting set of 28 features is provided in Table 1 of the supplemental material. The surrogate features are expected to change temporally as the surrogate model is updated after every surrogate optimisation cycle --- and indeed, the population is undergoing evolution and the algorithm is moving through different search space regions. Next to the surrogate landscape features, the true landscape features are extracted after every surrogate optimisation cycle. Subsequently, all solutions evaluated with the true fitness function, up to the current surrogate optimisation cycle, are used to quantify the optimisation convergence of the optimisation run. Optimisation convergence is evaluated by first normalising the objectives using the ideal and Nadir points, as recommended in the literature \cite{brockhoff2022using}. The Nadir point then corresponds to $[1, 1]$. In the context of multi-objective optimisation, we are only interested in regions where there is a trade-off between objectives, which are the solutions between the ideal and Nadir point. The median landscape feature and performance metric is calculated over 15 repeats.

\subsection{Data representation for analyses}
For every problem and surrogate model pair, data is composed of median surrogate landscape features, true landscape features, and a performance metric as a time-series. For computational expense purposes, only the data at the time instances (true function evaluations) $256$, $1280$, $1792$, and $8192$ are considered.


\section{Experimental setup} 
\label{ExperimentalSetup}
\subsection{Benchmark problems}
The bi-objective BBOB functions (\textsc{bbob-biobj}) \cite{brockhoff2022using} with continuous problem variables from the COCO framework \cite{hansen2021coco} are considered. The 55 problems are a pair-wise combination of 10 well-understood single objective BBOB functions. In this work, we are only considering the first instance of each of the 55 \textsc{bbob-biobj} problems. We focus on the 20-dimensional instances only. All algorithms have their populations initialized in the default COCO ranges: $-100$ to $100$. The benchmark problems are simulated to be computationally expensive by setting the maximum number of true function evaluations to $8192$. The maximum runtime of each optimisation is set to 24 hours. The resulting output data\footnote{\url{{https://zenodo.org/records/10575290}}} and scripts\footnote{\url{https://zenodo.org/records/10496995}} can be found in public repositories.

\subsection{SA-EA Algorithms}
\paragraph{SA-EA variants and the settings}
For this work, we are considering six different surrogates. We are considering \emph{IDW}, \emph{IDWR}, and \emph{LR-KNN} as described in Section \ref{bg:models}. \emph{IDW} and \emph{IDWR} do not require any settings, and are implemented as described in Section \ref{bg:models}. For \emph{LR-KNN}, the $k$ is set to $32$ to capture the local slope of the fitness landscape. In this work, we also consider simply using k-nearest neighbours (\emph{KNN}), where the surrogate fitness is equivalent to the nearest solution in the set of solutions evaluated using the true fitness function. This surrogate is interesting to consider because this surrogate is fast to evaluate and does not interpolate between true evaluated solutions. Furthermore, many state-of-the-art SA-EA use Kriging. Therefore, this work also uses the Kriging surrogate (\emph{Kriging}) from the well-known K-RVEA \cite{chugh2016surrogate} as implemented in the PlatEMO \cite{PlatEMO}. Finally, a surrogate named \emph{No structure} is included. This surrogate randomly selects a solution from the true evaluated solutions and returns the fitness values for \textit{both} objectives. This surrogate functions as a baseline since it returns fitness values in the ranges of previous evaluated solutions, but there is no correlation between the true and surrogate fitness values.\

All SA-EA variants are initialized with $32$ solutions which are sampled using LHS. For all variants, the number of selected solutions per surrogate optimisation cycle, $\mu$, is set to $1$ so that the surrogate is updated as frequently as possible (every new true evaluation). In this work, the MATLAB/Octave code implementation of RVEA from PlatEMO is utilized \cite{PlatEMO}. RVEA is initialized with a population size of $32$ which is sampled using LHS. Furthermore, $\alpha$ is set to $1 \cdot 10^6$; this is intended to maximize convergence due to the low number of evaluations.

\subsection{Dimensionality reduction}
We seek to compare vectors of landscape features between true and surrogate fitness functions, and also between different surrogates. To this end, we use t-distributed stochastic neighbour embedding (t-SNE); this is a dimensionality reduction technique widely employed to visualize high-dimensional data in a lower-dimensional space while preserving the pairwise similarities between data points \cite{van2008visualizing}. Employing dimensionality reduction individually for each sample within a time-series sequence (such as the different points in evolution which we track in this study) could introduce unwarranted variability in the consecutive projections. This complicates the process of tracking the optimisation. Therefore, also, a dynamic t-SNE \cite{rauber2016visualizing} has been used in this work. For the experiments, we use a perplexity of $50$, sigma optimisation iterations of $50$, epoch count of $10000$, and we set the random seed to $1$. In the case of a dynamic t-SNE, the movement penality is set to $0.001$.

\subsection{Performance Modelling}
The modelling is conducted with random forest regression as the learning algorithm; it is used with its default hyperparameters, namely: 500 trees, with \(\frac{1}{3}N_f\) features included per split (where $N_f$ is the number of features); sampling \emph{with} replacement; the sample size is the same as the number of observations. The 28 features are computed using both the true fitness landscape and the surrogate, which leads to 56 features in the candidate pool. Four points in evolution are considered for the temporal approach: after 256 evaluations; after 1280; after 1782; and 8192. 

For the static analysis, all features are based on the true fitness function and there is a single snapshot of the 28 features. The sample size for the temporal and static analysis is the same: 2000 (100$D$) points. The response variable is the median (normalised using the ideal and Nadir points) final hypervolume obtained by the surrogate-assisted RVEA variants. The IDW-assisted RVEA is excluded from the algorithm performance analysis: it achieved non-zero hypervolume on only one out of 55 functions, making it impossible to build a meaningful model. 

We carry out recursive feature elimination (RFE)\footnote{\url{https://www.rdocumentation.org/packages/caret/versions/6.0-92/topics/rfe}}, bootstrapped for 1000 iterations. In accordance with the one-in-ten guideline \cite{harrell1984regression} for the ratio between features and observations, we limit the number of selected features to a maximum of five (there are 55 benchmark problems under study). Models are built using the features which were identified during RFE. To try and account for the potential effect of having a limited number of observations (the model could be susceptible to randomness in the data split), we bootstrap the models for 1000 iterations using a 80-20 training-validation split on the dataset. To quantify model performance, we consider the pseudo \(R^2\); this is computed as \(1-\frac{MSE}{variance(y)}\) where $y$ is the target variable (normalised final hypervolume achieved by the surrogate-assisted RVEA). If a model is very poor, the pseudo \(R^2\) can sometimes be less than zero; to preserve its meaning as the proportion of variance explained, we replace any negative values with zero. We report the bootstrap mean, median, and standard error of the model \(R^2\) on validation data.

\section{Results}\label{results} 
\subsection{Static compared to temporal FLA}
\begin{figure}
    \centering
    \includegraphics[width=0.99\linewidth]{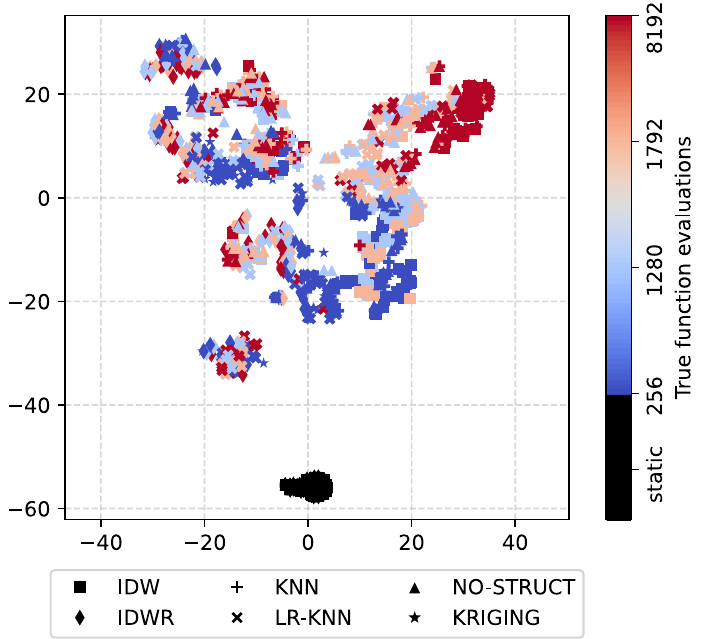}
    \caption{t-SNE plot comparing the true features resulting from the temporal FLA to the static FLA. Each marker represents the median fitness landscape feature of a particular \textsc{bbob-biobj} problem. The colour of the point indicates in what phase of the evolution the landscape feature is calculated, as indicated with the colour bar.}
    \label{fig:static-tSNE-dynamic-vs-static-using-true-features}
    \vspace{-1.5em}
\end{figure}

To ascertain whether the temporal FLA yields distinct features in comparison to a static FLA, we compared the true landscape features for both analyses in a t-SNE plot. The results are depicted in Figure \ref{fig:static-tSNE-dynamic-vs-static-using-true-features}. When using the \emph{Kriging} surrogate, only 33 out of 55 \textsc{bbob-biobj} surpassed 256 function evaluations before the optimisation runtime limit, see Table 1 in the supplementary material. Only these 33 problems are included in the further analysis. Each marker, in Figure \ref{fig:static-tSNE-dynamic-vs-static-using-true-features}, represents true fitness landscape features of one of the \textsc{bbob-biobj} problems. The black markers denote the features resulting from the static FLA, while the coloured markers indicate the phase of the evolution in the temporal FLA. 

The most significant finding here is the distinct discrepancy between markers derived from static and temporal FLA, indicating that the two types of analysis exhibit divergent feature distributions. Moreover, the feature similarity between the different \textsc{bbob-biobj} problems appears to be larger than the similarity between the features at different stages of evolution (in particular, notice the red cluster in the top-left of the plot containing the majority of the late-in-evolution samples), emphasizing the need to perform fitness landscape analyses temporally. We have also evaluated, per feature, whether there is a statistical difference between the static and temporal analysis (see supplementary material).

\subsection{True compared to surrogate landscapes}
\begin{figure}
    \centering
    \includegraphics[width=0.8\linewidth]{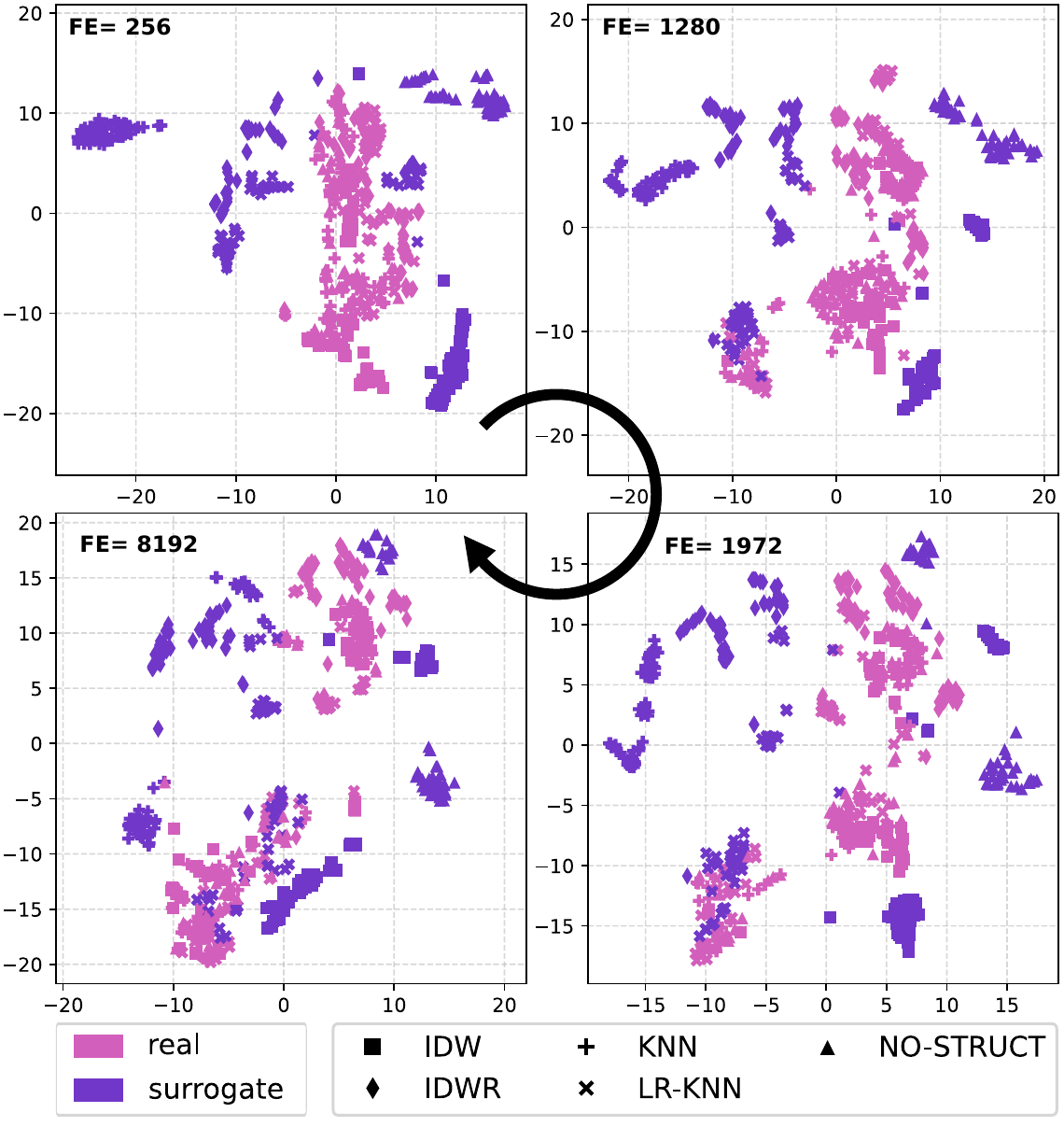}
    \caption{Dynamic t-SNE plot comparing fitness landscape features for the true to the surrogate fitness evaluations acquired in the temporal FLA. Each marker in the scatter plot represents the median fitness landscape feature of a particular \textsc{bbob-biobj} problem}
    \label{fig:dynamic-tSNE-surrugate-vs-true-features}
    \setlength{\belowcaptionskip}{-0.5\baselineskip}
    \vspace{-1.5em}
\end{figure}

We also compare the true landscape features \emph{and} the surrogate landscape features during different phases of evolution, see Figure \ref{fig:dynamic-tSNE-surrugate-vs-true-features}. The top-left plot represents the features at the beginning of the optimisation (256 function evaluations). Following the plots clockwise, there are different phases in evolution, as indicated with the annotated text. In the plots, each marker represents fitness landscape features of one of the \textsc{bbob-biobj} problems. The colours now describe whether the features are based on the true or surrogate landscape. The surrogate (indicated using the marker shapes) is also relevant for the true landscape features, since different surrogates can lead the optimisation to different regions of the true fitness landscape. \emph{Kriging} has been excluded for the analyses because we only have the data until 256 function evaluations. After 256 function evaluations, it is interesting that all true-landscape fitness features cluster together (notice the pink markers) while the surrogate features (these are coloured purple) disperse into the peripheral regions of the plot; this suggests a large difference between the true and the surrogate features. At the same time, the surrogate model data points are locally clustered together (into their surrogate types), suggesting there is also a clear difference between the features of different surrogates. 
To investigate the similarity between the surrogate and the underlying true landscape, we compare the corresponding feature distributions with a statistical test. Since in Figure \ref{fig:dynamic-tSNE-surrugate-vs-true-features}, the different surrogates form individual clusters throughout the evolution, the statistical comparison is performed per surrogate --- resulting in 588 tests. Because the surrogate landscape features are calculated on the same solutions as the true landscape features, a Wilcoxon test is used; we consider statistical significance to be $p \leq$ 0.05 (corrected using the Bonferroni correction).   

\begin{figure*}
    \centering
    \includegraphics[width=1\linewidth]{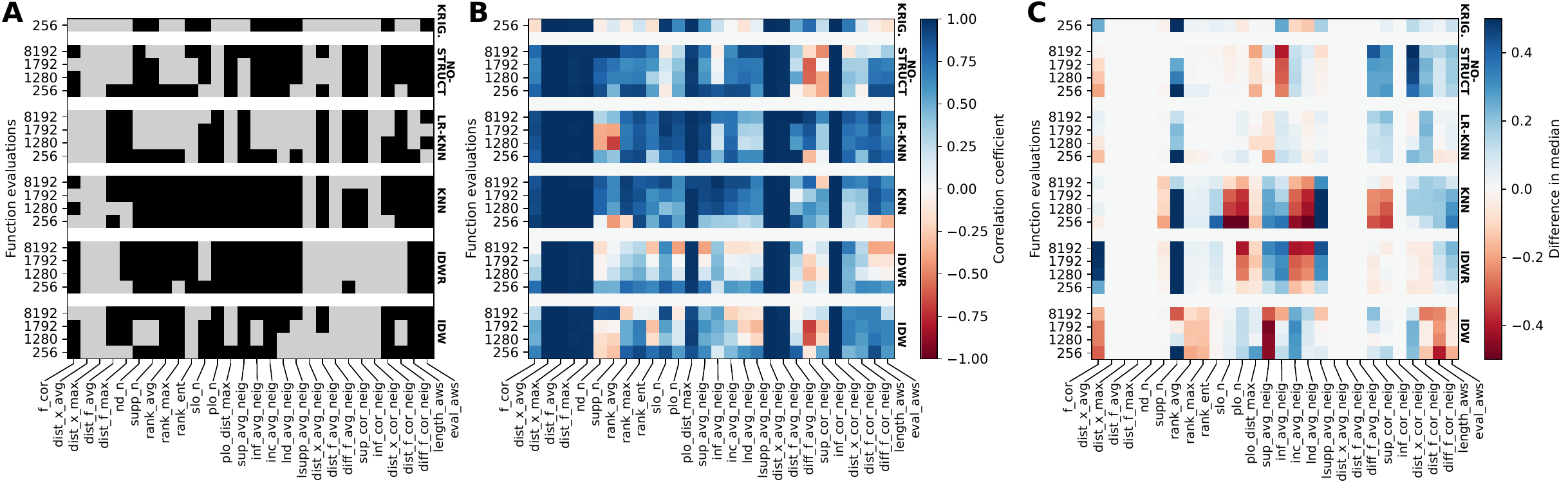}
    \caption{A) Comparing the true and surrogate landscape feature distributions yielded during different phases of evolution, where black refers to statistically significant ($p$ $\leq$ 0.05) differences according to a Wilcoxon test. B) Correlations between the true and surrogate landscape feature distributions yielded during different phases of evolution. C) Difference in median between the true and surrogate landscape feature distributions yielded during different phases of evolution.}
    \label{fig:stat-cor-med}
\end{figure*}

Figure \ref{fig:stat-cor-med}(a) summarises the results, where the black squares represent a significant difference in distributions. For \emph{KNN}, 17 of the 28 surrogate features are significantly different from the true features throughout the evolution; for \emph{IDWR}, \emph{No structure}, \emph{LR-KNN}, and \emph{IDW} this is: 14, 10, 8, and 8 features respectively. Furthermore, \emph{LR-KNN} seems to capture the true landscape from 1280 function evaluations: this can be observed in the increase in light-grey squares when going up the vertical axis towards 8192 evaluations. This does not occur for \emph{IDWR} and \emph{KNN}. Interestingly, despite these discrepancies, when we consider the Spearman correlations between the true and surrogate landscape feature distributions, the features do seem to correlate with each other: see Figure \ref{fig:stat-cor-med}(b). Overall, indicated by the number of blue squares, \emph{No structure}, \emph{LR-KNN}, and \emph{KNN} seem to have higher feature correlation than \emph{IDWR} and \emph{IDW}. Note that, for \emph{KNN}, there are many features whose distributions showed significant difference in \ref{fig:stat-cor-med}(a) while simultaneously having high correlations. We also notice that several features appear to have perfect correlations between the true and surrogate distributions. In the case of \emph{dist-x-avg}, \emph{dist-x-avg-neig}, and \emph{dist-x-max} --- this is because these relate to distance in the variable space. For the others (\emph{dist-f-avg}, \emph{dist-f-max}, and \emph{dist-f-avg-neig}): these three have highly skewed distributions for both true and surrogate landscapes, with the vast majority of values being low but with a few very large outliers. These distributions likely occurred due to the difference in fitness ranges across functions, and the nature of the distributions is probably the reason for the strong correlations. This raises the thought about whether this specific subset of features should perhaps be revisited or further developed in future research. 

To understand the distributions further we also visualise the difference between the true-landscape median and the surrogate-landscape median of the (normalised) feature distributions: see Figure \ref{fig:stat-cor-med}(c). For some features, particularly for \emph{KNN} and \emph{IDWR}, there is a large difference in the median of the feature distributions, explaining the significant differences seen in Figure \ref{fig:stat-cor-med}(a).

\begin{table*}[ht!]
\centering
\caption{Algorithm performance prediction models. The predictors are true (\emph{black text}) and surrogate (\emph{\textcolor{blue}{blue text}}) landscape features sampled temporally at four points in evolution (and a static sample) chosen by feature selection; the response variable is final (normalised) hypervolume achieved by the surrogate-assisted RVEA variant indicated in the first column. Predictors and the response are the median over 15 runs. The model quality metric is pseudo $R^2$. Each model is bootstrapped for 1000 iterations and the reported values are \emph{bootstrap mean} | \emph{bootstrap median} (\emph{bootstrap standard error}) for the metric on validation data.}
\vspace{2mm}
\resizebox{0.90\textwidth}{!}{\begin{tabular}{l|ccccc}
\toprule
sampling $\rightarrow$ & temporal (256 evals) & temporal (1280) & temporal (1792) & temporal (8192) & static \\ [0.2cm]
\midrule
 selected  $\rightarrow$ &  \begin{tabular}{c}[\emph{\textcolor{blue}{dist-f-max}, \textcolor{blue}{dist-f-avg}}, \\ \emph{dist-f-max, \textcolor{blue}{dist-f-avg-neig}},\\ \emph{\textcolor{blue}{diff-f-avg-neig}}] \end{tabular}
 &  \begin{tabular}{c}[\emph{\textcolor{blue}{inf-avg-neig}, slo-n,} \\ \emph{diff-f-avg-neig,} \\ \emph{\textcolor{blue}{dist-f-avg-neig, dist-f-avg}}] \end{tabular}&\begin{tabular}{c}[\emph{\textcolor{blue}{inf-avg-neig}, plo-dist-max,} \\ \emph{\textcolor{blue}{dist-x-avg-neig}, dist-x-avg-neig,} \\ \emph{\textcolor{blue}{nd-n}}] \end{tabular} & \begin{tabular}{c}[\emph{\textcolor{blue}{diff-f-cor-neig, dist-f-cor-neig,}} \\ \emph{\textcolor{blue}{lsupp-avg-neig}}, \\ \emph{diff-f-cor-neig, rank-avg}] \end{tabular} & \begin{tabular}{c}[\emph{dist-f-cor-neig, dist-f-avg-neig,} \\ \emph{dist-f-max, dist-f-avg,} \\ \emph{diff-f-avg-neig}] \end{tabular} \\ [0.2cm]
\emph{KNN} &  $R^2$: [0.755 | 0.753 (0.236)] & $R^2$: [0.819 | 0.855 (0.176)] & $R^2$: [0.913 | 0.858 (0.174)] & $R^2$: [0.947 | 0.934 (0.096)] & $R^2$: [0.697 | 0.785 (0.209)]  \\[0.2cm]
\midrule
selected $\rightarrow$ &  \begin{tabular}{c}[\emph{sup-avg-neig, eval-aws,} \\ \emph{inc-avg-neig, dist-f-avg-neig,} \\ \emph{\textcolor{blue}{diff-f-avg-neig}}] \end{tabular} &  \begin{tabular}{c}[\emph{\textcolor{blue}{diff-f-cor-neig, length-aws}}, \\ \emph{sup-avg-neig, \textcolor{blue}{dist-x-cor-neig}}, \\ \emph{dist-x-cor-neig}] \end{tabular} & \begin{tabular}{c}[\emph{\textcolor{blue}{diff-f-cor-neig}, sup-avg-neig,} \\ \emph{dist-x-cor-neig,} \\ \emph{\textcolor{blue}{dist-x-cor-neig, length-aws}}]\end{tabular} &  \begin{tabular}{c}[\emph{slo-n, nd-n,} \\ \emph{\textcolor{blue}{dist-f-cor-neig}, supp-n,} \\ \emph{\textcolor{blue}{diff-f-cor-neig}}] \end{tabular}&\begin{tabular}{c}[\emph{dist-f-cor-neig, sup-cor-neig,} \\ \emph{dist-f-avg, dist-f-avg-neig,} \\ \emph{diff-f-cor-neig}] \end{tabular}  \\ [0.2cm]
\emph{IDW} &  $R^2$: [0.682 | 0.698 (0.223)] & $R^2$: [0.834 | 0.786 (0.201)] & $R^2$: [0.748 | 0.763 (0.204)] & $R^2$: [0.870 | 0.866 (0.181)]  &  $R^2$: [0.754 | 0.749 (0.192)] \\[0.2cm]
\midrule
selected $\rightarrow$ &  \begin{tabular}{c}[\emph{\textcolor{blue}{plo-n, rank-ent}}, \\ \emph{\textcolor{blue}{plo-dist-max}, plo-dist-max, dist-f-avg}] \end{tabular} &  \begin{tabular}{c}[\emph{\textcolor{blue}{plo-dist-max, slo-n}}, \\ \emph{\textcolor{blue}{plo-n}, plo-dist-max, slo-n}] \end{tabular} & \begin{tabular}{c}[\emph{\textcolor{blue}{slo-n}, sup-avg-neig}, \\ \emph{\textcolor{blue}{plo-dist-max, plo-n}, slo-n}]\end{tabular} &  \begin{tabular}{c}[\emph{nd-n, plo-dist-max,} \\ \emph{\textcolor{blue}{slo-n}, inf-avg-neig,} \\ \emph{slo-n}] \end{tabular} & \begin{tabular}{c}[\emph{dist-f-cor-neig, dist-f-avg-neig,} \\ \emph{dist-f-avg, diff-f-avg-neig,} \\ \emph{dist-f-max}] \end{tabular} \\ [0.2cm]
\emph{LR-KNN} &  $R^2$: [0.867 | 0.832 (0.163)] & $R^2$: [0.967 | 0.837 (0.181)] & $R^2$: [0.945 | 0.846 (0.180)] & $R^2$: [0.923 | 0.836 (0.185)]  & $R^2$: [0.871 | 0.754 (0.213)]  \\[0.2cm]
\midrule
selected $\rightarrow$ &  \begin{tabular}{c}[\emph{\textcolor{blue}{diff-f-cor-neig, diff-f-avg-neig}}, \\ \emph{\textcolor{blue}{dist-f-cor-neig},} \\ \emph{dist-f-avg-neig, dist-f-avg}] \end{tabular} &  \begin{tabular}{c}[\emph{slo-n, \textcolor{blue}{dist-x-avg-neig}}, \\ \emph{dist-x-avg-neig,} \\ \emph{\textcolor{blue}{rank-ent, dist-f-cor-neig}}] \end{tabular} & \begin{tabular}{c}[\emph{dist-x-avg-neig, \textcolor{blue}{dist-x-avg-neig}}, \\ \emph{\textcolor{blue}{dist-x-avg}, dist-x-avg, slo-n}]\end{tabular} &  \begin{tabular}{c}[\emph{nd-n, \textcolor{blue}{nd-n}}, \\ \emph{\textcolor{blue}{supp-n}, supp-n,} \\ \emph{\textcolor{blue}{eval-aws}}] \end{tabular} & \begin{tabular}{c}[\emph{dist-f-cor-neig, dist-f-avg-neig,} \\ \emph{sup-cor-neig,} \\ \emph{dist-f-avg, diff-f-avg-neig}] \end{tabular} \\ [0.2cm]
 \emph{No structure}  &  $R^2$: [0.686 | 0.730 (0.215)] & $R^2$: [0.815 | 0.786 (0.171)]  & $R^2$: [0.778 | 0.754 (0.202)]  & $R^2$: [0.618 | 0.838 (0.172)]   & $R^2$: [0.728 | 0.701 (0.223)] \\[0.2cm]
\end{tabular}}
\label{tab:regression-models}
\vspace{-1em}
\end{table*}

\subsection{Algorithm performance prediction}

Table \ref{tab:regression-models} presents a summary of the algorithm performance prediction models. Each row relates to a specific type of surrogate, and contains information about five models (one per column), where the response variable is the final hypervolume associated with using that surrogate. For example, the first row represents: four models using temporal features from sampling during runs using the KNN-surrogate, and one model utilising \emph{static} features extracted from a latin hypercube sample. The type of sampling is indicated in the header row; for each row below that, \emph{selected} reports the features included in the model (these were selected by RFE) --- features of the surrogate landscape are in \emph{\textcolor{blue}{blue text}} and features of the true landscape are in \emph{black text}; and beside the name of the surrogate we present the bootstrap mean, median, and standard error for the model pseudo \(R^2\) on validation data. The \(R^2\) values should not be taken as an indication of the performance of the surrogates, but as an indicator of how well landscape features can predict the outcome of optimisation. 

Looking at the \(R^2\) values in Table \ref{tab:regression-models} we can consider the values for the models built using a static landscape sample as the baseline for interpretation. For the \emph{KNN} and \emph{LRKNN} surrogates, three of the four temporal models are better than the static equivalent; the remaining one is the first in the time series (256 evaluations). For \emph{IDW}, there are two superior temporal models, one which is roughly equal to the static equivalent, and one which is lower quality than the static counterpart (this is again the earliest temporal sample, at 256 evaluations). Finally, for the \emph{No structure} surrogate, it is not clear whether temporal or static analysis is the best approach: some of the temporal models are better then the static model and some are not. For both \emph{IDW} and \emph{No structure} surrogates, the models at 256 evaluations are worse than the static baseline. This is probably because the sampling, which is steered by surrogate fitnesses, is led to less relevant regions: this behaviour could be due to the limited archive of real-evaluated solutions at this point in the search and the poor ability of these two surrogates (indeed, they exhibit lower performance on the problems than \emph{KNN} and \emph{LRKNN}).

Notice from Table \ref{tab:regression-models} that for each of the 16 models built using temporal features, a mixture of both surrogate and true landscape features were selected by the RFE. This hints that it is important to consider features of both the surrogate \textbf{and} the true landscape when considering performance prediction for surrogate-assisted optimisation; however, to test this thought, we also built models using feature pools of 1) surrogate features only and 2) true features only. This experiment showed that generally, the surrogate-only and true-only models are of approximately equivalent performance to those in Table \ref{tab:regression-models}. That finding implies that, firstly, the surrogate models may be mimicking the true landscape well; and secondly, that features of either can be used in performance prediction. The results for these two setups can be found in Tables 2 and 3 of the supplemental material. In Table \ref{tab:regression-models}, it can be seen that fitness-based features (those containing \emph{f} in either blue or black text) appear to play an integral role in many of the predictions: the majority of models has at least one included. 

When it comes to these fitness-based features, more from the surrogate landscapes are selected when compared to the true landscapes. We notice that the type of feature which matters appears to change over the phases of evolution: when the 1792 timepoint is reached, there is a shift away from fitness-based features towards features which capture evolvability and ruggedness with respect to variable-space distances (those ending with \emph{avg-neig} and \emph{cor-neig}). Looking again at the temporal models overall, there is also a clear trend of features relating to local optima --- both single-objective and Pareto --- appearing in the models (for example: \emph{slo-n}, \emph{plo-n}, and \emph{plo-dist-max}). Notice that this is particularly evident in the case of the \emph{LRKNN} surrogate, and that these features seem to be important regardless of whether they are computed from the surrogate landscape or the true landscape. The prevalence of local optima features in the \emph{LRKNN} models implies that the performance of this type of surrogate may be particularly affected by local optima. 

In terms of model quality over time, for the \emph{KNN} surrogate there is a trend over time: the model quality increases. There is no clear trend for the other three surrogates. As mentioned, the last column contains relating to the models using features of the static sample as predictors. We observe that the majority of selected features for these models relate to fitness in some way. Comparing the static models with the temporal equivalents, it can be seen that all four of the temporal \emph{KNN} models out-perform the static \emph{KNN} model; and three of the \emph{LR-KNN} temporal models outperform the equivalent static model. 

Things are less clear with the \emph{IDW} and \emph{No structure} surrogates. Taking into account the four temporal models, it could be said that the equivalent static models perform similarly to them. It may seem counter-intuitive that the \emph{No structure} models have mean \(R^2\) values between 0.618 and 0.815. This surrogate simply assigns a random true-evaluated fitness to a solution. However, when we consider the details of these models, things become clearer. For example: in the case of the temporal (256 evaluations) model, the surrogate does not have access to many true-evaluated solutions yet so the sample for landscape analysis --- and subsequently the predictive model --- resembles that of the static sample in column 5. For the second and third temporal models, a lot of features based on distance in the variable space are selected (these begin with \emph{dist-x}); this makes sense, because they are not directly related to the surrogate fitness function. The feature \emph{\textcolor{blue}{dist-f-cor-neig}} (surrogate) is selected for the 1280-evaluations model. This is a fitness-based feature (and therefore highly linked to the surrogate), but the fact it was selected for the models is understandable when we consider that this feature showed moderate correlation with its equivalent for the true landscape, as seen in Figure \ref{fig:stat-cor-med}b). To conclude this section, we note that feature importances were computed for all models, and that generally, features within a model have similar importance to one another. The associated data can be found in Tables 3, 5, and 7 of the supplemental material. 

 \subsection{Limitations and Discussion}

 The performance prediction results indicate that temporal fitness landscape analysis potentially be used for online surrogate selection in multi-objective optimisation. This ratifies what has been shown in a single-objective context \cite{le2013evolution}. We also found that while many landscape features significantly differ between the surrogate and true landscapes, they are often correlated. A previous study \cite{werth2020surrogate} observed feature differences for single-objective problems. Additionally, our results showed that using either or both of the surrogate and true landscape features can be beneficial in performance prediction; to the best of our knowledge, this is a novel approach: previous studies have focused on the evolvability of the surrogate \cite{le2013evolution} or the fitness approximation error \cite{bagheri2016online}. 
 
There are limitations to the conducted study. 
The feature selection approach --- recursive feature elimination --- can be affected by randomness in the data splits, and can sometimes fail to remove redundant features. Even so, every feature selection method has its own drawbacks: for example, evolutionary feature selection is computationally intense and may require parameter tuning. We also note that sampling strategies for conducting temporal landscape analysis introduce additional computational expense. Future work will consider how to best utilise the fitness evaluations which have already been carried out during optimisation. Another consideration is that the included surrogate models were chosen due to their straightforwardness; however, in future works, more Kriging models from other state-of-the-art SA-EAs such as in \cite{zhang2009expensive, chugh2016surrogate} and other types of surrogate models such as neural networks in \cite{guo2021evolutionary} should be considered. Finally, some of these models can become expensive to build and evaluate when the archive of solutions becomes too large. The impact of the fitness landscapes when reducing the archive size should also be considered.

\section{Conclusions} 
\label{sec:conclusions}
Surrogate-assisted evolutionary algorithms can effectively exploit problem structures by substituting expensive-to-evaluate fitness functions. Analysing the resemblance and discrepancies between the surrogate landscape and the true landscape could provide a better understanding how surrogates can assist EAs in performing efficient optimisation by capturing various features of the true fitness landscape. In this work, we have investigated the difference between true and surrogate landscape features during evolutionary multi-objective optimisation. We consider several surrogate-assisted versions of a reference-vector guided evolutionary algorithm and use the well-known \textsc{BBOB-BIOBJ} suite of bi-objective functions. Our results indicate that surrogate landscape features differ significantly from the true landscape features and that these features vary during the course of a run. Despite these differences, the surrogate and true landscape often show a high correlation. This work also evaluates how these different features impact the actual search and identified key landscape features from both the surrogate and the true landscape with the capability to predict algorithm performance. This opens the door for online surrogate switching in multi-objective optimisation in the future. 

\section*{Acknowledgments}
This research is part of the research programme Open Technology Programme with project number 15586, which is financed by the Dutch Research Council (NWO), Elekta, and Xomnia. Further, the work is co-funded by the public-private partnership allowance for top consortia for knowledge and innovation (TKIs) from the Dutch Ministry of Economic Affairs.

\clearpage

\appendix

\section{Fitness landscape features}

\begin{table}[h]
\small
\centering
\caption{Description of the considered landscape features, all of which are from \cite{liefooghe2021landscape}.}
\begin{tabular}{l|l}
\textbf{Features}           & Description \\ \hline
\textbf{f\_cor}             & correlation among objective values  \\
\textbf{dist\_x\_avg}       & average distance among solutions in the parameter space \\
\textbf{dist\_x\_max}       & maximum distance among solutions in the parameter space  \\
\textbf{dist\_f\_avg}       & average distance among solutions in the objective space \\
\textbf{dist\_f\_max}       & maximum distance among solutions in the objective space  \\
\textbf{nd\_n}              & proportion of non-dominated solutions \\
\textbf{supp\_n}            & proportion of supported non-dominated solutions \\
\textbf{rank\_avg}          & average rank w.r.t. non-dominated sorting \\
\textbf{rank\_max}          & maximum rank w.r.t. non-dominated sorting  \\
\textbf{rank\_ent}          & entropy of the number of solutions per rank \\
\textbf{slo\_n}             & proportion of single-objective local optima per objective \\
\textbf{plo\_n}             & proportion of Pareto local optima \\
\textbf{plo\_dist\_max}     & avg. distance among Pareto local optima in the var. space \\
\textbf{sup\_avg\_neig}     & avg. proportion of dominating neighbours  \\
\textbf{inf\_avg\_neig}     & avg. proportion of dominated neighbours  \\
\textbf{inc\_avg\_neig}     & avg. proportion of incomparable neighbours \\
\textbf{lnd\_avg\_neig}     & avg. proportion of locally non-dominated neighbours \\
\textbf{lsupp\_avg\_neig}   & avg. proportion of supported locally non-dominated neig. \\
\textbf{dist\_x\_avg\_neig} & avg. distance from neighbours in the parameter space \\
\textbf{dist\_f\_avg\_neig} & avg. distance from neighbours in the objective space \\
\textbf{diff\_f\_avg\_neig} & avg. difference per objective with neig. \\
\textbf{sup\_cor\_neig}     & neig.’s cor. of the proportion of dominating neig. \\
\textbf{inf\_cor\_neig}     & neig.’s cor. of the proportion of dominated neig. \\
\textbf{dist\_x\_cor\_neig} & neig.’s cor. of the avg. distance from neig. in the var. space \\
\textbf{dist\_f\_cor\_neig} & neig.’s cor. of the avg. distance from neig. in the obj. space \\
\textbf{diff\_f\_cor\_neig} & neig.’s cor. of the avg. difference per objective from neig. \\
\textbf{length\_aws}        & average length of adaptive walks      \\
\textbf{eval\_aws}          & adaptive walk evaluations \\
\bottomrule
\end{tabular}
\label{tab:features}
\vspace{-1em}
\end{table}

\section{Experimental setup details}
\subsection{Hardware specifications}
The optimisation and feature extraction process is executed using high-performance computing composed out of AMD EPYC 7H12 nodes, each containing 64 cores with a base clock speed of 2.6GHz and 2 GB of memory per core. The runtime of each independent run in the optimisation process is limited to 24 hours. The optimisation process uses Octave version 7.3.0 and the COCO framework \cite{hansen2021coco}\footnote{\url{https://github.com/numbbo/coco}} cloned in 2023. The feature extraction process uses \textsf{R} 4.2.1 and scripts which are made publicly available alongside the paper which proposed the metrics \cite{liefooghe2021landscape}\footnote{\url{https://gitlab.com/aliefooghe/landscape-features-mo-icops}} cloned in 2023. The resulting output data\footnote{\url{{https://zenodo.org/records/10575290}}} and scripts\footnote{\url{Empty}} can be found in public repositories. The dimensionality reduction is executed in an Ubuntu 22.04 environment using a Lenovo IdeaPad 5 Pro machine containing 8 AMD Ryzen 7 6800HS cores with a clock speed of 3,2GHz, 16 GB of memory and NVIDIA GeForce RTX 3050 Ti GPU. The dynamic t-SNE is from \cite{rauber2016visualizing}\footnote{\url{https://github.com/paulorauber/thesne}} cloned in 2023. The performance modelling is executed using \textsf{R} in an Ubuntu 22.04 environment on a machine with 20 12th Gen Intel i9-12900HK cores and 32GB of memory.

\begin{table}[]
    \centering
        \caption{For each function, the number of independent runs using $Kriging$ which surpassed 256 true function evaluations within the time limit of the optimisation process (24 hours).}
    \begin{tabular}{|c c||c c||c c||c c|} \hline
    \textbf{Func.}  & \textbf{Runs} & \textbf{Func.} & \textbf{Runs} & \textbf{Func.} & \textbf{Runs} & \textbf{Func.} & \textbf{Runs} \\ \hline
    1  & 0  & 15 & 2 & 29 & 0  & 43 & 2 \\ \hline
    2  & 6  & 16 & 1 & 30 & 0  & 44 & 0 \\ \hline
    3  & 3  & 17 & 1 & 31 & 0  & 45 & 12\\ \hline
    4  & 0  & 18 & 4 & 32 & 0  & 46 & 14\\ \hline
    5  & 0  & 19 & 6 & 33 & 0  & 47 & 15\\ \hline
    6  & 9  & 20 & 3 & 34 & 0  & 48 & 11\\ \hline
    7  & 0  & 21 & 0 & 35 & 3  & 49 & 14\\ \hline
    8  & 0  & 22 & 5 & 36 & 8  & 50 & 13\\ \hline
    9  & 0  & 23 & 3 & 37 & 2  & 51 & 13\\ \hline
    10 & 0  & 24 & 1 & 38 & 1  & 52 & 12\\ \hline
    11 & 11 & 25 & 0 & 39 & 0  & 53 & 0 \\ \hline
    12 & 9  & 26 & 1 & 40 & 0  & 54 & 0 \\ \hline
    13 & 6  & 27 & 3 & 41 & 12 & 55 & 0 \\ \hline
    14 & 5  & 28 & 0 & 42 & 3  &    &   \\ \hline
    \end{tabular}
    \label{tab:my_label}
\end{table}

\begin{figure}[h!]
    \centering
    \includegraphics[width=0.9\linewidth]{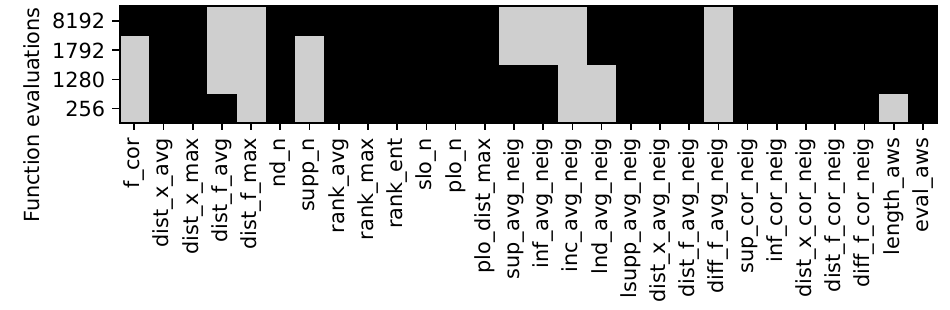}
    \caption{Comparing the individual landscape feature distributions yielded from the static FLA to the feature distributions yielded from temporal FLA during different phases of evolution; where black refers to statistically significant according to a Mann-Whitney U test and and $p$-value $\leq$ 0.05.}
    \label{fig:blok-dynamic-vs-static-using-true-individual-feature-distributions}
\end{figure}

\section{Static compared to temporal analysis}
To investigate the differences between the static and dynamic FLA of all 28 features individually, the true landscape feature distributions are split by sampling type and are compared with a statistical test at four phases of the evolution. This results in 112 Mann-Whitney U tests; we consider a $p$-value of 0.05 or less (corrected using the Bonferroni correction) as an indication of statistical significance. Because the \emph{Kriging} surrogate does not include runs for each \textsc{bbob-biobj} problem and time points, this surrogate is excluded from the rest of this work. In Figure \ref{fig:blok-dynamic-vs-static-using-true-individual-feature-distributions}, all black squares denote that there is a significant difference between the distributions for a feature of the temporal and static FLA. The figure shows that 18 out of 28 features have significantly different static and temporal distributions regardless of which timepoint the temporal features come from, and only for three out of 28 features all four temporal timepoints have no indication of statistical difference with the static distributions. 

\section{Performance Prediction Models}
Tables \ref{tab:regression-models}, \ref{tab:regression-models-onlysurr}, and \ref{tab:regression-models-onlyreal} show performance prediction results for models using \emph{both} surrogate and true landscape features, \emph{only} surrogate features, and \emph{only} true features, respectively. The associated feature importance estimates are provided in Tables \ref{tab:regression-models-importances}, \ref{tab:regression-models-onlysurr-importances}, and \ref{tab:regression-models-onlyreal-importances}. 

\begin{table*}[hb!]
\centering
\caption{Algorithm performance prediction models. The predictors are true (\emph{black text}) and surrogate (\emph{\textcolor{blue}{blue text}}) landscape features sampled temporally at four points in evolution (and a static sample) chosen by feature selection; the response variable is final (normalised) hypervolume achieved by the surrogate-assisted RVEA variant indicated in the first column. Predictors and the response are the median over 15 runs. The model quality metric is pseudo $R^2$. Each model is bootstrapped for 1000 iterations and the reported values are \emph{bootstrap mean} | \emph{bootstrap median} (\emph{bootstrap standard error}) for the metric on validation data. Feature importances are provided in Table 3.}
\vspace{2mm}
\resizebox{0.99\textwidth}{!}{\begin{tabular}{l|ccccc}
\toprule
sampling $\rightarrow$ & temporal (256 evals) & temporal (1280) & temporal (1792) & temporal (8192) & static \\ [0.2cm]
\midrule
 selected  $\rightarrow$ &  \begin{tabular}{c}[\emph{\textcolor{blue}{dist-f-max}, \textcolor{blue}{dist-f-avg}}, \\ \emph{dist-f-max, \textcolor{blue}{dist-f-avg-neig}},\\ \emph{\textcolor{blue}{diff-f-avg-neig}}] \end{tabular}
 &  \begin{tabular}{c}[\emph{\textcolor{blue}{inf-avg-neig}, slo-n,} \\ \emph{diff-f-avg-neig,} \\ \emph{\textcolor{blue}{dist-f-avg-neig, dist-f-avg}}] \end{tabular}&\begin{tabular}{c}[\emph{\textcolor{blue}{inf-avg-neig}, plo-dist-max,} \\ \emph{\textcolor{blue}{dist-x-avg-neig}, dist-x-avg-neig,} \\ \emph{\textcolor{blue}{nd-n}}] \end{tabular} & \begin{tabular}{c}[\emph{\textcolor{blue}{diff-f-cor-neig, dist-f-cor-neig,}} \\ \emph{\textcolor{blue}{lsupp-avg-neig}}, \\ \emph{diff-f-cor-neig, rank-avg}] \end{tabular} & \begin{tabular}{c}[\emph{dist-f-cor-neig, dist-f-avg-neig,} \\ \emph{dist-f-max, dist-f-avg,} \\ \emph{diff-f-avg-neig}] \end{tabular} \\ [0.2cm]
\emph{KNN} &  $R^2$: [0.755 | 0.753 (0.236)] & $R^2$: [0.819 | 0.855 (0.176)] & $R^2$: [0.913 | 0.858 (0.174)] & $R^2$: [0.947 | 0.934 (0.096)] & $R^2$: [0.697 | 0.785 (0.209)]  \\[0.2cm]
\midrule
selected $\rightarrow$ &  \begin{tabular}{c}[\emph{sup-avg-neig, eval-aws,} \\ \emph{inc-avg-neig, dist-f-avg-neig,} \\ \emph{\textcolor{blue}{diff-f-avg-neig}}] \end{tabular} &  \begin{tabular}{c}[\emph{\textcolor{blue}{diff-f-cor-neig, length-aws}}, \\ \emph{sup-avg-neig, \textcolor{blue}{dist-x-cor-neig}}, \\ \emph{dist-x-cor-neig}] \end{tabular} & \begin{tabular}{c}[\emph{\textcolor{blue}{diff-f-cor-neig}, sup-avg-neig,} \\ \emph{dist-x-cor-neig,} \\ \emph{\textcolor{blue}{dist-x-cor-neig, length-aws}}]\end{tabular} &  \begin{tabular}{c}[\emph{slo-n, nd-n,} \\ \emph{\textcolor{blue}{dist-f-cor-neig}, supp-n,} \\ \emph{\textcolor{blue}{diff-f-cor-neig}}] \end{tabular}&\begin{tabular}{c}[\emph{dist-f-cor-neig, sup-cor-neig,} \\ \emph{dist-f-avg, dist-f-avg-neig,} \\ \emph{diff-f-cor-neig}] \end{tabular}  \\ [0.2cm]
\emph{IDW} &  $R^2$: [0.682 | 0.698 (0.223)] & $R^2$: [0.834 | 0.786 (0.201)] & $R^2$: [0.748 | 0.763 (0.204)] & $R^2$: [0.870 | 0.866 (0.181)]  &  $R^2$: [0.754 | 0.749 (0.192)] \\[0.2cm]
\midrule
selected $\rightarrow$ &  \begin{tabular}{c}[\emph{\textcolor{blue}{plo-n, rank-ent}}, \\ \emph{\textcolor{blue}{plo-dist-max}, plo-dist-max, dist-f-avg}] \end{tabular} &  \begin{tabular}{c}[\emph{\textcolor{blue}{plo-dist-max, slo-n}}, \\ \emph{\textcolor{blue}{plo-n}, plo-dist-max, slo-n}] \end{tabular} & \begin{tabular}{c}[\emph{\textcolor{blue}{slo-n}, sup-avg-neig}, \\ \emph{\textcolor{blue}{plo-dist-max, plo-n}, slo-n}]\end{tabular} &  \begin{tabular}{c}[\emph{nd-n, plo-dist-max,} \\ \emph{\textcolor{blue}{slo-n}, inf-avg-neig,} \\ \emph{slo-n}] \end{tabular} & \begin{tabular}{c}[\emph{dist-f-cor-neig, dist-f-avg-neig,} \\ \emph{dist-f-avg, diff-f-avg-neig,} \\ \emph{dist-f-max}] \end{tabular} \\ [0.2cm]
\emph{LR-KNN} &  $R^2$: [0.867 | 0.832 (0.163)] & $R^2$: [0.967 | 0.837 (0.181)] & $R^2$: [0.945 | 0.846 (0.180)] & $R^2$: [0.923 | 0.836 (0.185)]  & $R^2$: [0.871 | 0.754 (0.213)]  \\[0.2cm]
\midrule
selected $\rightarrow$ &  \begin{tabular}{c}[\emph{\textcolor{blue}{diff-f-cor-neig, diff-f-avg-neig}}, \\ \emph{\textcolor{blue}{dist-f-cor-neig},} \\ \emph{dist-f-avg-neig, dist-f-avg}] \end{tabular} &  \begin{tabular}{c}[\emph{slo-n, \textcolor{blue}{dist-x-avg-neig}}, \\ \emph{dist-x-avg-neig,} \\ \emph{\textcolor{blue}{rank-ent, dist-f-cor-neig}}] \end{tabular} & \begin{tabular}{c}[\emph{dist-x-avg-neig, \textcolor{blue}{dist-x-avg-neig}}, \\ \emph{\textcolor{blue}{dist-x-avg}, dist-x-avg, slo-n}]\end{tabular} &  \begin{tabular}{c}[\emph{nd-n, \textcolor{blue}{nd-n}}, \\ \emph{\textcolor{blue}{supp-n}, supp-n,} \\ \emph{\textcolor{blue}{eval-aws}}] \end{tabular} & \begin{tabular}{c}[\emph{dist-f-cor-neig, dist-f-avg-neig,} \\ \emph{sup-cor-neig,} \\ \emph{dist-f-avg, diff-f-avg-neig}] \end{tabular} \\ [0.2cm]
 \emph{No structure}  &  $R^2$: [0.686 | 0.730 (0.215)] & $R^2$: [0.815 | 0.786 (0.171)]  & $R^2$: [0.778 | 0.754 (0.202)]  & $R^2$: [0.618 | 0.838 (0.172)]   & $R^2$: [0.728 | 0.701 (0.223)] \\[0.2cm]
\end{tabular}}
\label{tab:regression-models}
\end{table*}

\begin{table*}[hb!]
\centering
\caption{Feature importances for the performance prediction models in Table 2. Importance is the mean (over 500 trees) decrease in \emph{node impurity}, totalled across every split which used the feature. The node impurity is quantified as the residual sum of squares. Importance values reported are the median over 1000 bootstrapping iterations.}
\vspace{2mm}
\resizebox{0.60\textwidth}{!}{\begin{tabular}{l|ccccc}
\toprule
sampling $\rightarrow$ & temporal (256) & temporal (1280) & temporal (1792) & temporal (8192) & static \\ [0.2cm]
\midrule
\emph{KNN}  $\rightarrow$ &  \begin{tabular}{c}[\emph{\textcolor{blue}{1.288}, \textcolor{blue}{1.336}}, \\ \emph{1.287, \textcolor{blue}{1.303}},\\ \emph{\textcolor{blue}{1.276}}] \end{tabular}
 &  \begin{tabular}{c}[\emph{\textcolor{blue}{1.422}, 1.230,} \\ \emph{1.306,} \\ \emph{\textcolor{blue}{1.289, 1.289}}] \end{tabular}&\begin{tabular}{c}[\emph{\textcolor{blue}{1.333}, 1.320,} \\ \emph{\textcolor{blue}{1.314}, 1.322,} \\ \emph{\textcolor{blue}{1.308}}] \end{tabular} & \begin{tabular}{c}[\emph{\textcolor{blue}{0.702, 1.704,}} \\ \emph{\textcolor{blue}{1.658}}, \\ \emph{1.012, 1.561}] \end{tabular} & \begin{tabular}{c}[\emph{1.429, 1.279,} \\ \emph{1.276, 1.280,} \\ \emph{1.262}] \end{tabular} \\ [0.2cm] \midrule
\emph{IDW} $\rightarrow$ &  \begin{tabular}{c}[\emph{1.057, 1.039,} \\ \emph{1.056, 1.098,} \\ \emph{\textcolor{blue}{1.095}}] \end{tabular} &  \begin{tabular}{c}[\emph{\textcolor{blue}{1.199, 1.132}}, \\ \emph{1.022, \textcolor{blue}{1.037}}, \\ \emph{1.038}] \end{tabular} & \begin{tabular}{c}[\emph{\textcolor{blue}{1.103}, 1.109,} \\ \emph{1.094,} \\ \emph{\textcolor{blue}{1.091, 1.043}}]\end{tabular} &  \begin{tabular}{c}[\emph{1.138, 1.148,} \\ \emph{\textcolor{blue}{1.095}, 1.091,} \\ \emph{\textcolor{blue}{1.057}}] \end{tabular}&\begin{tabular}{c}[\emph{1.272, 1.117,} \\ \emph{1.096, 1.087,} \\ \emph{0.872}] \end{tabular}  \\ [0.2cm]
\midrule
\emph{LRKNN} $\rightarrow$ &  \begin{tabular}{c}[\emph{\textcolor{blue}{1.383, 0.948}}, \\ \emph{\textcolor{blue}{1.409}, 1.422, 1.387}] \end{tabular} &  \begin{tabular}{c}[\emph{\textcolor{blue}{1.356, 1.331}}, \\ \emph{\textcolor{blue}{1.332}, 1.332, 1.234}] \end{tabular} & \begin{tabular}{c}[\emph{\textcolor{blue}{1.345}, 1.329}, \\ \emph{\textcolor{blue}{1.326, 1.301}, 1.284}]\end{tabular} &  \begin{tabular}{c}[\emph{1.337, 1.334,} \\ \emph{\textcolor{blue}{1.336}, 1.279,} \\ \emph{1.289}] \end{tabular} & \begin{tabular}{c}[\emph{1.570, 1.233,} \\ \emph{1.237, 1.231,} \\ \emph{1.227}] \end{tabular} \\ [0.2cm]
\midrule
\emph{No structure} $\rightarrow$ &  \begin{tabular}{c}[\emph{\textcolor{blue}{1.192, 0.987}}, \\ \emph{\textcolor{blue}{1.196},} \\ \emph{1.183, 1.181}] \end{tabular} &  \begin{tabular}{c}[\emph{1.283, \textcolor{blue}{1.262}}, \\ \emph{1.262,} \\ \emph{\textcolor{blue}{0.897, 1.106}}] \end{tabular} & \begin{tabular}{c}[\emph{1.224, \textcolor{blue}{1.218}}, \\ \emph{\textcolor{blue}{1.172}, 1.171, 1.03}]\end{tabular} &  \begin{tabular}{c}[\emph{1.203, \textcolor{blue}{1.179}}, \\ \emph{\textcolor{blue}{1.180}, 1.183,} \\ \emph{\textcolor{blue}{1.112}}] \end{tabular} & \begin{tabular}{c}[\emph{1.329, 1.158,} \\ \emph{0.955,} \\ \emph{1.163, 1.136}] \end{tabular} \\ [0.2cm]
\end{tabular}}
\label{tab:regression-models-importances}
\end{table*}

\begin{table*}[ht!]
\centering
\caption{Algorithm performance prediction models using \textbf{only} features of the \emph{surrogate} landscape. The predictors are landscape features sampled temporally at four points in evolution (and from a static sample) chosen by feature selection; the response variable is final (normalised) hypervolume achieved by the surrogate-assisted RVEA variant indicated in the first column. Predictors and the response are the median over 15 runs. The model quality metric is pseudo $R^2$. Each model is bootstrapped for 1000 iterations and the reported values are \emph{bootstrap mean} | \emph{bootstrap median} (\emph{bootstrap standard error}) for the metric on validation data. Feature importances are provided in Table 5.}
\vspace{2mm}
\resizebox{0.99\textwidth}{!}{\begin{tabular}{l|ccccc}
\toprule
sampling $\rightarrow$ & temporal (256 evals) & temporal (1280) & temporal (1792) & temporal (8192) & static \\ [0.2cm]
\midrule
 selected  $\rightarrow$ &  \begin{tabular}{c}[\emph{\textcolor{blue}{diff-f-cor-neig, dist-f-cor-neig}}, \\ \emph{\textcolor{blue}{lsupp-avg-neig, dist-x-cor-neig,}} \\  \emph{\textcolor{blue}{dist-x-avg-neig}}] \end{tabular}
 &  \begin{tabular}{c}[\textcolor{blue}{\emph{inf-avg-neig, dist-f-avg-neig,}} \\ \emph{\textcolor{blue}{dist-f-avg, dist-f-max,}} \\ \emph{\textcolor{blue}{lnd-avg-neig}}] \end{tabular}&\begin{tabular}{c}[\emph{\textcolor{blue}{dist-x-avg-neig, inf-avg-neig,}} \\ \emph{\textcolor{blue}{nd-n, dist-x-avg,}} \\ \emph{\textcolor{blue}{dist-f-max}}] \end{tabular} & \begin{tabular}{c}[\emph{\textcolor{blue}{diff-f-cor-neig, dist-f-cor-neig,}} \\ \emph{\textcolor{blue}{lsupp-avg-neig, dist-x-cor-neig,}} \\ \emph{\textcolor{blue}{dist-x-avg-neig}}] \end{tabular}  & \begin{tabular}{c}[\emph{dist-f-cor-neig, dist-f-avg-neig,} \\ \emph{dist-f-max, dist-f-avg,} \\ \emph{diff-f-avg-neig}] \end{tabular} \\ [0.2cm]
\emph{KNN} &  $R^2$: [0.379 | 0.517 (0.224)] & $R^2$: [0.774 | 0.835 (0.190)] & $R^2$: [0.884 | 0.867 (0.177)] & $R^2$: [0.960 | 0.938 (0.092)] & $R^2$: [0.697 | 0.785 (0.209)]  \\[0.2cm]
\midrule
selected $\rightarrow$ &  \begin{tabular}{c}[\emph{\textcolor{blue}{dist-f-avg, dist-f-avg-neig,}} \\ \emph{\textcolor{blue}{diff-f-avg-neig, dist-f-max,}} \\ \emph{\textcolor{blue}{sup-avg-neig}}] \end{tabular} &  \begin{tabular}{c}[\emph{\textcolor{blue}{diff-f-cor-neig, length-aws,}} \\ \emph{\textcolor{blue}{dist-x-cor-neig, dist-f-max,}} \\ \emph{\textcolor{blue}{dist-f-avg}}] \end{tabular} & \begin{tabular}{c}[\textcolor{blue}{\emph{dist-x-cor-neig, diff-f-cor-neig,}} \\ \emph{\textcolor{blue}{length-aws, rank-avg,}} \\ \emph{\textcolor{blue}{dist-f-avg}}]\end{tabular} &  \begin{tabular}{c}[\textcolor{blue}{\emph{dist-f-cor-neig, diff-f-cor-neig,}} \\ \emph{\textcolor{blue}{dist-x-avg-neig, slo-n,}} \\ \emph{\textcolor{blue}{eval-aws}}] \end{tabular} &\begin{tabular}{c}[\emph{dist-f-cor-neig, sup-cor-neig,} \\ \emph{dist-f-avg, dist-f-avg-neig,} \\ \emph{diff-f-cor-neig}] \end{tabular}  \\ [0.2cm]
\emph{IDW} &  $R^2$: [0.803 | 0.703 (0.231)] & $R^2$: [0.805 | 0.788 (0.195)] & $R^2$: [0.759 | 0.761 (0.211)]  &  $R^2$: [0.867 | 0.797 (0.194)]  &  $R^2$: [0.754 | 0.749 (0.192)] \\[0.2cm]
\midrule
selected $\rightarrow$ &  \begin{tabular}{c}[\emph{\textcolor{blue}{plo-n, plo-dist-max,}} \\ \emph{\textcolor{blue}{rank-ent, dist-f-avg,}} \\ \emph{\textcolor{blue}{dist-f-avg-neig}}] \end{tabular} &  \begin{tabular}{c}[\emph{\textcolor{blue}{plo-dist-max, slo-n,}} \\ \emph{\textcolor{blue}{plo-n, inc-avg-neig,}} \\ \emph{\textcolor{blue}{length-aws}}] \end{tabular} & \begin{tabular}{c}[\textcolor{blue}{\emph{plo-dist-max, slo-n,}} \\ \emph{\textcolor{blue}{plo-n, sup-avg-neig,}} \\ \emph{\textcolor{blue}{nd-n}}]\end{tabular} &  \begin{tabular}{c}[\textcolor{blue}{\emph{slo-n, plo-dist-max,}} \\ \emph{\textcolor{blue}{dist-x-max, dist-x-avg-neig,}} \\ \emph{\textcolor{blue}{diff-f-cor-neig}}] \end{tabular} & \begin{tabular}{c}[\emph{dist-f-cor-neig, dist-f-avg-neig,} \\ \emph{dist-f-avg, diff-f-avg-neig,} \\ \emph{dist-f-max}] \end{tabular} \\ [0.2cm]
\emph{LR-KNN} &  $R^2$: [0.879 | 0.818 (0.163)] & $R^2$: [0.945 | 0.831 (0.182)] & $R^2$: [0.937 | 0.857 (0.165)]  & $R^2$: [0.946 | 0.834 (0.190)]   & $R^2$: [0.871 | 0.754 (0.213)]  \\[0.2cm]
\midrule
selected $\rightarrow$ & \begin{tabular}{c}[\emph{\textcolor{blue}{diff-f-cor-neig, diff-f-avg-neig,}} \\ \emph{\textcolor{blue}{dist-f-cor-neig, dist-f-max,}} \\ \emph{\textcolor{blue}{dist-f-avg}}] \end{tabular} &  \begin{tabular}{c}[\emph{\textcolor{blue}{dist-x-avg-neig, dist-x-max,}} \\ \emph{\textcolor{blue}{dist-f-cor-neig, dist-x-avg,}} \\ \emph{\textcolor{blue}{rank-ent}}] \end{tabular} & \begin{tabular}{c}[\emph{\textcolor{blue}{dist-x-avg-neig, dist-x-avg,}} \\ \emph{\textcolor{blue}{dist-x-max, plo-dist-max,}} \\ \emph{\textcolor{blue}{diff-f-cor-neig}}]\end{tabular} &  \begin{tabular}{c}[\emph{\textcolor{blue}{nd-n, supp-n,}} \\ \emph{\textcolor{blue}{eval-aws, dist-x-cor-neig,}} \\ \emph{\textcolor{blue}{dist-x-avg-neig}}] \end{tabular} & \begin{tabular}{c}[\emph{dist-f-cor-neig, dist-f-avg-neig,} \\ \emph{sup-cor-neig,} \\ \emph{dist-f-avg, diff-f-avg-neig}] \end{tabular} \\ [0.2cm]
 \emph{No structure}  &  $R^2$: [0.684 | 0.733 (0.213)] & $R^2$: [0.745 | 0.774 (0.197)]  & $R^2$: [0.751 | 0.779 (0.198)]   & $R^2$: [0.630 | 0.817 (0.186)]  & $R^2$: [0.728 | 0.701 (0.223)] \\[0.2cm]
\bottomrule
\end{tabular}}
\label{tab:regression-models-onlysurr}
\end{table*}

\begin{table*}[ht!]
\centering
\caption{Feature importances for the performance prediction models in Table 4. Importance is the mean (over 500 trees) decrease in \emph{node impurity}, totalled across every split which used the feature. The node impurity is quantified as the residual sum of squares. Importance values reported are the median over 1000 bootstrapping iterations.}
\vspace{2mm}
\resizebox{0.6\textwidth}{!}{\begin{tabular}{l|ccccc}
\toprule
sampling $\rightarrow$ & temporal (256) & temporal (1280) & temporal (1792) & temporal (8192) & static \\ [0.2cm]
\midrule
\emph{KNN}  $\rightarrow$ &  \begin{tabular}{c}[\emph{\textcolor{blue}{1.081, 1.072}}, \\ \emph{\textcolor{blue}{1.394, 1.166,}} \\  \emph{\textcolor{blue}{1.447}}] \end{tabular}
 &  \begin{tabular}{c}[\textcolor{blue}{\emph{1.444, 1.328,}} \\ \emph{\textcolor{blue}{1.321, 1.309,}} \\ \emph{\textcolor{blue}{1.167}}] \end{tabular}&\begin{tabular}{c}[\emph{\textcolor{blue}{1.353, 1.371,}} \\ \emph{\textcolor{blue}{1.332, 1.294,}} \\ \emph{\textcolor{blue}{1.235}}] \end{tabular} & \begin{tabular}{c}[\emph{\textcolor{blue}{0.599, 1.570}}, \\ \emph{\textcolor{blue}{1.514, 1.472, 1.515}} \end{tabular} & \begin{tabular}{c}[\emph{1.429, 1.279,} \\ \emph{1.276, 1.280,} \\ \emph{1.262}] \end{tabular} \\ [0.2cm] 
\midrule
\emph{IDW} $\rightarrow$ &  \begin{tabular}{c}[\emph{\textcolor{blue}{1.102, 1.092,}} \\ \emph{\textcolor{blue}{1.094, 1.094,}} \\ \emph{\textcolor{blue}{1.047}}] \end{tabular} &  \begin{tabular}{c}[\emph{\textcolor{blue}{1.277, 1.203,}} \\ \emph{\textcolor{blue}{1.137, 0.917,}} \\ \emph{\textcolor{blue}{0.928}}] \end{tabular} & \begin{tabular}{c}[\textcolor{blue}{\emph{1.198, 1.167,}} \\ \emph{\textcolor{blue}{1.106, 1.056,}} \\ \emph{\textcolor{blue}{0.940}}]\end{tabular} &  \begin{tabular}{c}[\textcolor{blue}{\emph{1.210, 1.180,}} \\ \emph{\textcolor{blue}{1.031, 0.968,}} \\ \emph{\textcolor{blue}{1.028}}] \end{tabular} & \begin{tabular}{c}[\emph{1.272, 1.117,} \\ \emph{1.096, 1.087,} \\ \emph{0.872}] \end{tabular}  \\ [0.2cm]
\midrule
\emph{LRKNN} $\rightarrow$ &  \begin{tabular}{c}[\emph{\textcolor{blue}{1.418, 1.422,}} \\ \emph{\textcolor{blue}{0.961, 1.374,}} \\ \emph{\textcolor{blue}{1.361}}] \end{tabular} &  \begin{tabular}{c}[\emph{\textcolor{blue}{1.424, 1.382,}} \\ \emph{\textcolor{blue}{1.379, 1.219,}} \\ \emph{\textcolor{blue}{1.134}}] \end{tabular} & \begin{tabular}{c}[\textcolor{blue}{\emph{1.386, 1.395,}} \\ \emph{\textcolor{blue}{1.362, 1.302,}} \\ \emph{\textcolor{blue}{1.107}}]\end{tabular} &  \begin{tabular}{c}[\textcolor{blue}{\emph{1.458, 1.392,}} \\ \emph{\textcolor{blue}{1.295, 1.137,}} \\ \emph{\textcolor{blue}{1.288}}] \end{tabular} & \begin{tabular}{c}[\emph{1.570, 1.233,} \\ \emph{1.237, 1.231,} \\ \emph{1.227}] \end{tabular} \\ [0.2cm]
\midrule
\emph{No structure} $\rightarrow$ & \begin{tabular}{c}[\emph{\textcolor{blue}{1.206, 1.009,}} \\ \emph{\textcolor{blue}{1.217, 1.162,}} \\ \emph{\textcolor{blue}{1.161}}] \end{tabular} &  \begin{tabular}{c}[\emph{\textcolor{blue}{1.296, 1.154,}} \\ \emph{\textcolor{blue}{1.161, 1.211,}} \\ \emph{\textcolor{blue}{0.955}}] \end{tabular} & \begin{tabular}{c}[\emph{\textcolor{blue}{1.280, 1.205,}} \\ \emph{\textcolor{blue}{1.200, 1.136,}} \\ \emph{\textcolor{blue}{1.001}}]\end{tabular} &  \begin{tabular}{c}[\emph{\textcolor{blue}{1.272, 1.273,}} \\ \emph{\textcolor{blue}{1.205, 0.948,}} \\ \emph{\textcolor{blue}{1.130}}] \end{tabular} & \begin{tabular}{c}[\emph{1.329, 1.158,} \\ \emph{0.955,} \\ \emph{1.163, 1.136}] \end{tabular} \\ [0.2cm]
\bottomrule
\end{tabular}}
\label{tab:regression-models-onlysurr-importances}
\end{table*}

\begin{table*}[ht!]
\centering
\caption{Algorithm performance prediction models using \textbf{only} features of the \emph{true} landscape. The predictors are landscape features sampled temporally at four points in evolution (and from a static sample) chosen by feature selection; the response variable is final (normalised) hypervolume achieved by the surrogate-assisted RVEA variant indicated in the first column. Predictors and the response are the median over 15 runs. The model quality metric is pseudo $R^2$. Each model is bootstrapped for 1000 iterations and the reported values are \emph{bootstrap mean} | \emph{bootstrap median} (\emph{bootstrap standard error}) for the metric on validation data. Feature importances are provided in Table 7.}
\vspace{2mm}
\resizebox{0.99\textwidth}{!}{\begin{tabular}{l|ccccc}
\toprule
sampling $\rightarrow$ & temporal (256 evals) & temporal (1280) & temporal (1792) & temporal (8192) & static \\ [0.2cm]
\midrule
 selected  $\rightarrow$ &  \begin{tabular}{c}[\emph{dist-f-max, diff-f-avg-neig,} \\ \emph{dist-f-avg, dist-f-avg-neig}] \end{tabular}
 &  \begin{tabular}{c}[\emph{diff-f-avg-neig, slo-n,} \\ \emph{dist-f-avg-neig, dist-f-max,} \\ \emph{dist-f-avg}] \end{tabular}& \begin{tabular}{c}[\emph{plo-dist-max, dist-x-avg-neig,} \\ \emph{dist-x-avg, sup-avg-neig,} \\ \emph{slo-n}] \end{tabular} & \begin{tabular}{c}[\emph{dist-x-cor-neig, diff-f-cor-neig,} \\ \emph{rank-avg, dist-x-avg-neig,} \\ \emph{dist-x-max}] \end{tabular}  & \begin{tabular}{c}[\emph{dist-f-cor-neig, dist-f-avg-neig,} \\ \emph{dist-f-max, dist-f-avg,} \\ \emph{diff-f-avg-neig}] \end{tabular} \\ [0.2cm]
\emph{KNN} &  $R^2$: [0.799 | 0.717 (0.254)] & $R^2$: [0.775 | 0.819 (0.196)] & $R^2$: [0.916 | 0.869 (0.159)] & $R^2$: [0.964 | 0.910 (0.125)] & $R^2$: [0.697 | 0.785 (0.209)]  \\[0.2cm]
\midrule
selected $\rightarrow$ &  \begin{tabular}{c}[\emph{dist-f-avg-neig, dist-f-avg,} \\ \emph{dist-f-max, sup-avg-neig,} \\ \emph{eval-aws}] \end{tabular} &  \begin{tabular}{c}[\emph{dist-x-cor-neig, sup-avg-neig,} \\ \emph{sup-cor-neig, inf-cor-neig,} \\ \emph{dist-f-max}] \end{tabular} & \begin{tabular}{c}[\emph{dist-x-cor-neig, sup-avg-neig,} \\ \emph{plo-dist-max, sup-cor-neig,} \\ \emph{plo-n}]\end{tabular} &  \begin{tabular}{c}[\emph{slo-n, nd-n,} \\ \emph{supp-n, dist-f-cor-neig,} \\ \emph{dist-x-avg-neig}] \end{tabular} &\begin{tabular}{c}[\emph{dist-f-cor-neig, sup-cor-neig,} \\ \emph{dist-f-avg, dist-f-avg-neig,} \\ \emph{diff-f-cor-neig}] \end{tabular}  \\ [0.2cm]
\emph{IDW} &  $R^2$: [0.650 | 0.690 (0.200)] & $R^2$: [0.769 | 0.756 (0.200)] & $R^2$: [0.761 | 0.817 (0.164)]  &  $R^2$: [0.913 | 0.852 (0.190)]  &  $R^2$: [0.754 | 0.749 (0.192)] \\[0.2cm]
\midrule
selected $\rightarrow$ &  \begin{tabular}{c}[\emph{plo-dist-max, dist-f-avg,} \\ \emph{dist-f-avg-neig, dist-f-max,} \\ \emph{diff-f-avg-neig}] \end{tabular} &  \begin{tabular}{c}[\emph{plo-dist-max, sup-avg-neig,} \\ \emph{slo-n, inf-avg-neig,} \\ \emph{plo-n}] \end{tabular} & \begin{tabular}{c}[\emph{sup-avg-neig, supp-n,} \\ \emph{slo-n, plo-dist-max,} \\ \emph{inf-avg-neig}]\end{tabular} &  \begin{tabular}{c}[\emph{nd-n, plo-dist-max,} \\ \emph{slo-n, supp-n,} \\ \emph{inf-avg-neig}] \end{tabular} & \begin{tabular}{c}[\emph{dist-f-cor-neig, dist-f-avg-neig,} \\ \emph{dist-f-avg, diff-f-avg-neig,} \\ \emph{dist-f-max}] \end{tabular} \\ [0.2cm]
\emph{LR-KNN} &  $R^2$: [0.799 | 0.762 (0.196)] & $R^2$: [0.945 | 0.815 (0.217)] & $R^2$: [0.902 | 0.837 (0.156)]  & $R^2$: [0.925 | 0.845 (0.194)]   & $R^2$: [0.871 | 0.754 (0.213)]  \\[0.2cm]
\midrule
selected $\rightarrow$ & \begin{tabular}{c}[\emph{dist-f-avg, dist-f-avg-neig,} \\ \emph{dist-f-max, diff-f-avg-neig,} \\ \emph{rank-ent}] \end{tabular} &  \begin{tabular}{c}[\emph{slo-n, dist-x-avg-neig,} \\ \emph{dist-x-max, dist-f-avg-neig,} \\ \emph{dist-x-avg}] \end{tabular} & \begin{tabular}{c}[\emph{dist-x-avg-neig, dist-x-avg,} \\ \emph{dist-x-max, plo-dist-max,} \\ \emph{slo-n}]\end{tabular} &  \begin{tabular}{c}[\emph{nd-n, supp-n,} \\ \emph{lsupp-avg-neig, dist-x-max,} \\ \emph{dist-x-cor-neig}] \end{tabular} & \begin{tabular}{c}[\emph{dist-f-cor-neig, dist-f-avg-neig,} \\ \emph{sup-cor-neig,} \\ \emph{dist-f-avg, diff-f-avg-neig}] \end{tabular} \\ [0.2cm]
 \emph{No structure}  &  $R^2$: [0.715 | 0.696 (0.235)] & $R^2$: [0.857 | 0.772 (0.180)]  & $R^2$: [0.792 | 0.757 (0.207)]   & $R^2$: [0.575 | 0.834 (0.184)]  & $R^2$: [0.728 | 0.701 (0.223)] \\[0.2cm]
\bottomrule
\end{tabular}}
\label{tab:regression-models-onlyreal}
\end{table*}

\begin{table*}[ht!]
\centering
\caption{Feature importances for the performance prediction models in Table 6. Importance is the mean (over 500 trees) decrease in \emph{node impurity}, totalled across every split which used the feature. The node impurity is quantified as the residual sum of squares. Importance values reported are the median over 1000 bootstrapping iterations.}
\vspace{2mm}
\resizebox{0.6\textwidth}{!}{\begin{tabular}{l|ccccc}
\toprule
sampling $\rightarrow$ & temporal (256) & temporal (1280) & temporal (1792) & temporal (8192) & static \\ [0.2cm]
\midrule
 \emph{KNN}  $\rightarrow$ &  \begin{tabular}{c}[\emph{1.638, 1.643,} \\ \emph{1.625, 1.626}] \end{tabular}
 &  \begin{tabular}{c}[\emph{1.346, 1.279,} \\ \emph{1.331, 1.304,} \\ \emph{1.300}] \end{tabular}& \begin{tabular}{c}[\emph{1.414, 1.409,} \\ \emph{1.341, 1.115,} \\ \emph{1.280}] \end{tabular} & \begin{tabular}{c}[\emph{1.422, 0.870,} \\ \emph{1.419, 1.473,} \\ \emph{1.427}] \end{tabular}  &  \begin{tabular}{c}[\emph{1.429, 1.279,} \\ \emph{1.276, 1.280,} \\ \emph{1.262}] \end{tabular} \\ [0.2cm] 
\midrule
\emph{IDW} $\rightarrow$ &  \begin{tabular}{c}[\emph{1.067, 1.058,} \\ \emph{1.080, 1.109,} \\ \emph{1.036}] \end{tabular} &  \begin{tabular}{c}[\emph{1.164, 1.144,} \\ \emph{1.099, 1.029} \\ \emph{0.958}] \end{tabular} & \begin{tabular}{c}[\emph{1.238, 1.237,} \\ \emph{1.058, 1.037,} \\ \emph{0.870}]\end{tabular} &  \begin{tabular}{c}[\emph{1.235, 1.253,} \\ \emph{1.203, 0.784,} \\ \emph{1.032}] \end{tabular} & \begin{tabular}{c}[\emph{1.272, 1.117,} \\ \emph{1.096, 1.087,} \\ \emph{0.872}] \end{tabular}  \\ [0.2cm]
\midrule
\emph{LRKNN} $\rightarrow$ &  \begin{tabular}{c}[\emph{1.396, 1.307,} \\ \emph{1.310, 1.253,} \\ \emph{1.247}] \end{tabular} &  \begin{tabular}{c}[\emph{1.433, 1.351,} \\ \emph{1.283, 1.242,} \\ \emph{1.219}] \end{tabular} & \begin{tabular}{c}[\emph{1.404, 1.216,} \\ \emph{1.348, 1.293,} \\ \emph{1.247}]\end{tabular} &  \begin{tabular}{c}[\emph{1.333, 1.341,} \\ \emph{1.298, 1.299,} \\ \emph{1.288}] \end{tabular} & \begin{tabular}{c}[\emph{1.570, 1.233,} \\ \emph{1.237, 1.231,} \\ \emph{1.227}] \end{tabular} \\ [0.2cm]
\midrule
\emph{No structure} $\rightarrow$ & \begin{tabular}{c}[\emph{1.236, 1.232,} \\ \emph{1.211, 1.221,} \\ \emph{0.843}] \end{tabular} &  \begin{tabular}{c}[\emph{1.256, 1.185,} \\ \emph{1.089, 1.128,} \\ \emph{1.122}] \end{tabular} & \begin{tabular}{c}[\emph{1.243, 1.188,} \\ \emph{1.172, 1.16,} \\ \emph{1.022}]\end{tabular} &  \begin{tabular}{c}[\emph{1.308, 1.266,} \\ \emph{1.227, 1.100,} \\ \emph{0.957}] \end{tabular} & \begin{tabular}{c}[\emph{1.329, 1.158,} \\ \emph{0.955,} \\ \emph{1.163, 1.136}] \end{tabular} \\ [0.2cm]
\bottomrule
\end{tabular}}
\label{tab:regression-models-onlyreal-importances}
\end{table*}

\clearpage
\bibliographystyle{ACM-Reference-Format}
\bibliography{sample-base}

\end{document}